\documentclass[sigconf]{acmart}
\usepackage{subfig}
\usepackage{multirow}
\usepackage{graphicx}
\usepackage{enumitem}
\AtBeginDocument{%
  }

\copyrightyear{2023}
\acmYear{2023}
\setcopyright{acmlicensed}\acmConference[CIKM '23]{Proceedings of the 32nd ACM International Conference on Information and Knowledge Management}{October 21--25, 2023}{Birmingham, United Kingdom}
\acmBooktitle{Proceedings of the 32nd ACM International Conference on Information and Knowledge Management (CIKM '23), October 21--25, 2023, Birmingham, United Kingdom}
\acmPrice{15.00}
\acmDOI{10.1145/3583780.3615249}
\acmISBN{979-8-4007-0124-5/23/10}

\begin{document}

\title{Class Label-aware Graph Anomaly Detection}


\author{Junghoon Kim}
\affiliation{%
  \institution{KAIST GSDS}
  \city{Daejeon}
  \country{Republic of Korea}}
\email{jhkim611@kaist.ac.kr}

\author{Yeonjun In}
\affiliation{%
  \institution{KAIST ISysE}
  \city{Daejeon}
  \country{Republic of Korea}}
\email{yeonjun.in@kaist.ac.kr}

\author{Kanghoon Yoon}
\affiliation{%
  \institution{KAIST ISysE}
  \city{Daejeon}
  \country{Republic of Korea}}
\email{ykhoon08@kaist.ac.kr}

\author{Junmo Lee}
\affiliation{%
  \institution{KAIST ISysE}
  \city{Daejeon}
  \country{Republic of Korea}}
\email{bubblego0217@kaist.ac.kr}

\author{Chanyoung Park}
\authornote{Corresponding author}
\affiliation{%
  \institution{KAIST ISysE \& AI}
  \city{Daejeon}
  \country{Republic of Korea}}
\email{cy.park@kaist.ac.kr}

\renewcommand{\shortauthors}{Kim et al.}

\begin{abstract}
  Unsupervised GAD methods assume the lack of anomaly labels, i.e., whether a node is anomalous or not. 
  One common observation we made from previous unsupervised methods is that they not only assume the absence of such anomaly labels, but also the \textit{absence of class labels} (the class a node belongs to used in a general node classification task). In this work, we study the utility of class labels for unsupervised GAD; in particular, how they enhance the detection of \textit{structural anomalies}. To this end, we propose a Class Label-aware Graph Anomaly Detection framework (CLAD) that utilizes a limited amount of labeled nodes to enhance the performance of unsupervised GAD. 
  Extensive experiments on ten datasets demonstrate the superior performance of CLAD in comparison to existing unsupervised GAD methods, even in the absence of ground-truth class label information. The source code for CLAD is available at \url{https://github.com/jhkim611/CLAD}.
\end{abstract}

\vspace{-1ex}

\vspace{-1ex}


\maketitle

\vspace{-1ex}
\section{Introduction}
\textit{Anomalies} in real-world graphs can manifest as
frauds in telecommunications networks~\cite{survey3, telecom}, fake reviews in user rating systems~\cite{yelp}, or illicit transactions in financial networks~\cite{elliptic}. Many Graph Anomaly Detection (GAD) studies employ Graph Neural Networks (GNNs) due to their superior performance on various downstream tasks~\cite{gcn, gat, prognn, rgcn}. 
A line of research for GAD adopts a supervised learning setting - where nodes are labeled as anomalous or benign - and solves a binary classification problem to detect anomalies~\cite{amnet, dagad, caregnn, pcgnn}. However, obtaining ground-truth anomaly labels is not an easy task. For many real-world cases, only a small fraction of anomalous samples are annotated as such~\cite{supervised_hard}, which makes the utilization of supervised methods in such scenarios comparatively cumbersome.

\begin{figure}[t]
\centering
\subfloat[]{
\includegraphics[height=2.2cm]{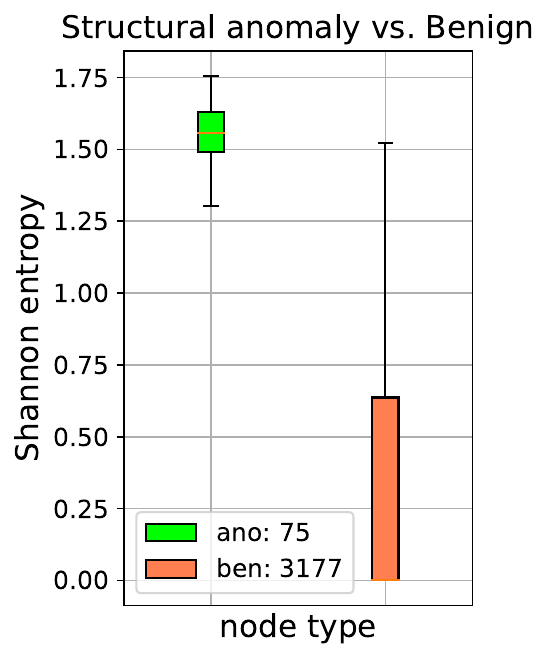}
\label{fig:entropy-a}
}
\subfloat[]{
\includegraphics[height=2.2cm]{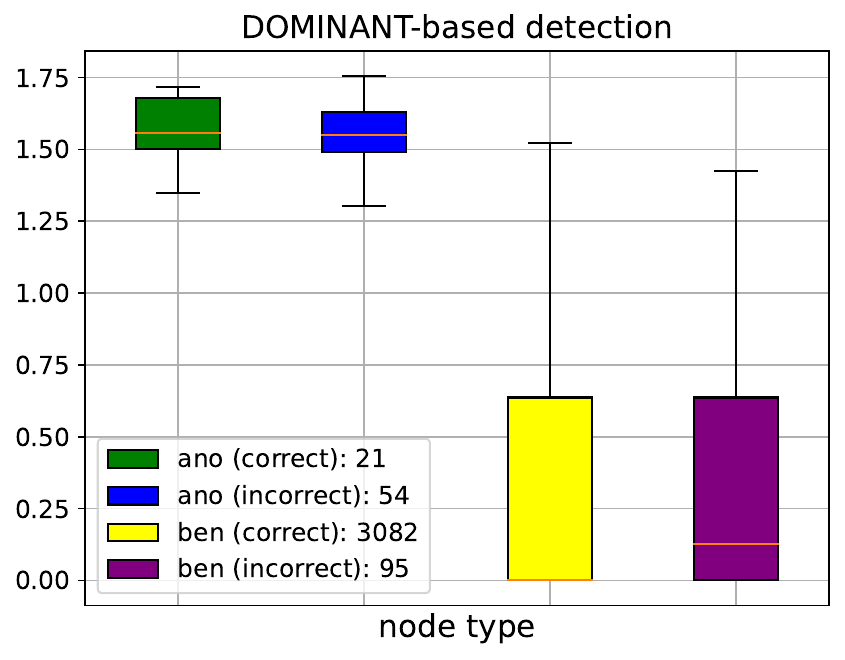}
\label{fig:entropy-b}
}
\subfloat[]{
\includegraphics[height=2.2cm]{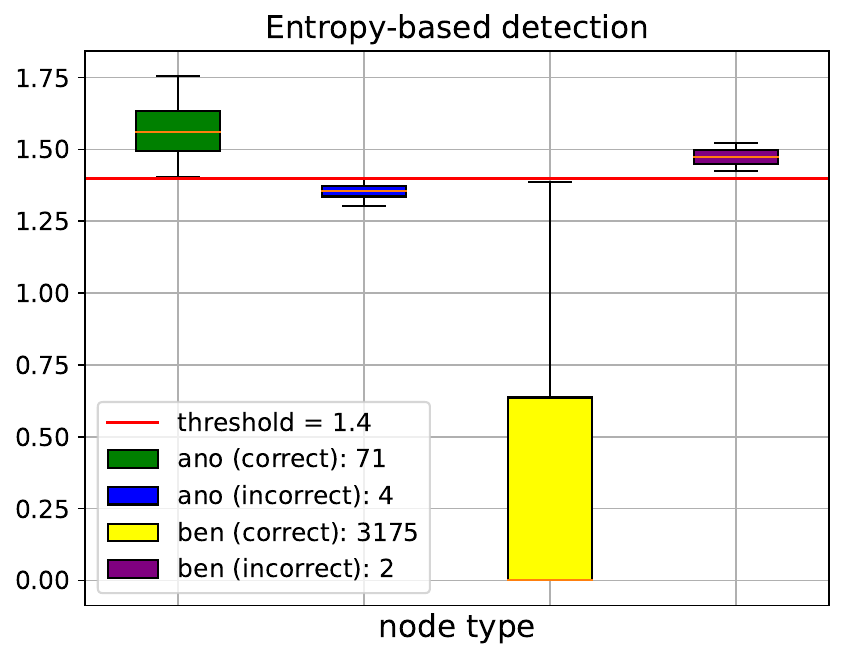}
\label{fig:entropy-c}
}
\vspace{-2ex}
\caption{(a) Average Shannon entropy of the neighborhood class distribution of structural anomalies/benign nodes. 
(b) GAD result of DOMINANT~\cite{dominant}. (c) GAD result of an entropy-based method, where the red horizontal line indicates the threshold (In (b) and (c), green/blue boxes indicate correctly/incorrectly classified structural anomalies, and yellow/purple boxes indicate correctly/incorrectly classified benign nodes. The number of nodes in each group is shown).}
\vspace{-6ex}
\label{fig:citeseer_entropy}
\end{figure}

On the other hand, unsupervised GAD methods assume the absence of such labels, which is a more realistic scenario. Previous studies~\cite{dominant, cola} categorize a node's malicious behaviors into \textit{structural anomalies} - nodes that exhibit abnormal connections with other nodes - and \textit{attribute anomalies} - nodes whose attributes significantly differ from those of their neighbors. 
Recent works~\cite{dominant, comga} have employed GNN-based autoencoders to assign anomaly scores to individual nodes based on reconstruction errors. Specifically, one GNN is trained to capture the structural information of nodes, while the other is trained to capture attribute information. 
The anomaly score for each node is computed as a weighted sum of the reconstruction errors from both GNNs, which do not require any anomaly label information.

One common observation we made from previous unsupervised methods is that they not only assume the absence of anomaly labels (whether node is anomalous used in an anomaly detection task), but also the \textit{absence of class labels} (the class a node belongs to used in a general node classification task). However, since a structural anomaly tends to exhibit abnormal behaviors in its vicinity (i.e., neighbors) in terms of class distribution, we find that the class label information of nodes is indeed beneficial for the GAD task. 
In Fig.~\ref{fig:entropy-a}, we show the average Shannon entropy of the neighborhood class distribution for both structural anomalies and benign nodes in Citeseer dataset. We clearly observe that the vast majority of benign nodes have a lower entropy compared with that of structural anomalies, indicating that structural anomalies tend to have neighboring nodes with a higher variety of classes. 
Based on this observation, in Fig.~\ref{fig:entropy-c}, we used the entropy value as a threshold (red line) to distinguish between structural anomalies and benign nodes. Surprisingly, we greatly outperformed the anomaly detection performance of an existing unsupervised GAD method, DOMINANT~\cite{dominant}, shown in Fig.~\ref{fig:entropy-b}, i.e., we reduced the number of anomalies that are incorrectly classified as benign (54$\rightarrow$4), and decreased the number of benign nodes that are incorrectly classified as anomalous (95$\rightarrow$2) as well. 

Inspired by our observations in Fig.~\ref{fig:citeseer_entropy}, we focus on the utility of class labels for unsupervised GAD. More specifically, we study how the class label information of nodes enhances the detection of \textit{structural anomalies} in particular.
In fact, the experiments in Fig.~\ref{fig:citeseer_entropy} are conducted assuming that the node labels are known for all nodes, which explains the superior performance of a simple entropy-based anomaly detection approach as shown in Fig.~\ref{fig:citeseer_entropy}c. 
However, since only a small fraction of nodes in real-world graphs are typically labeled, it becomes crucial to effectively utilize the limited amount of labeled nodes available. 

To this end, we propose a \underline{C}lass \underline{L}abel-aware Graph \underline{A}nomaly \underline{D}etection framework (CLAD) that utilizes a limited amount of labeled nodes to enhance the performance of unsupervised GAD. 
More precisely, the main component of CLAD is the structural anomaly quantifier that uses the output of a GNN-based node classifier to compute the discrepancy (i.e., Jenson-Shannon Divergence (JSD)) between the \textit{predicted} class distribution of a node and the average of its neighboring nodes. 

\looseness=-1
However, we discovered that the JSD value computed between a node and its neighboring nodes is highly dependent on the degree of the node. In other words, introducing an additional neighbor to a low-degree node would significantly impact the JSD value, whereas adding such a neighbor to a high-degree node would result in a negligible change. Hence, we propose a modified JSD, i.e., JSD+, to take into account the node degree information. Moreover, attribute anomalies are identified through an attribute anomaly quantifier, which is designed to compute the discrepancy between the attributes of a node and its neighboring nodes. In the end, we combine the anomaly scores obtained from the structural/attribute anomaly quantifiers to obtain the final anomaly score of each node.

Through extensive experiments, we demonstrate that CLAD outperforms existing unsupervised GAD methods, even when the ratio of known labels is extremely small. Moreover, even in an extreme case where no class labels are available at all, pseudo-labels constructed from simple attribute-based clustering techniques are enough for CLAD to successfully distinguish anomalies from benign nodes. 
This show that the framework is suitable for application even in scenarios where obtaining class labels is difficult. To the best of our knowledge, this is the first work to consider class labels of nodes for unsupervised GAD.

\section{Problem Statement}
\noindent{\textbf{Definition. }}
Let $\mathcal{G = (V, E, A, X)}$ denote an undirected attributed graph, where $\mathcal{V}$ is the set of $\mathit{N}$ nodes, $\mathcal{E}$ is the set of edges between these nodes and $\mathcal{X} \in \mathbb{R}^{\mathit{N \times F}}$ contains the node attributes. $\mathcal{A} \in \mathbb{R}^\mathit{N \times N}$ is the adjacency matrix, in which $\mathcal{A}_{\mathit{ij}} = 1$ indicates $\mathit{v_i}$ and $\mathit{v_j}$ are linked and $\mathcal{A}_{\mathit{ij}} = 0$ otherwise. $\mathit{F}$ is the number of node attributes and $\mathit{x_i} = \mathcal{X}[\mathit{i}, :] \in \mathbb{R}^{\mathit{F}}$ indicates the attribute vector of node $\mathit{v_i}$.

\smallskip
\noindent{\textbf{Problem. }}
Given an attributed graph $\mathcal{G = (V, E, A, X)}$, our goal is to detect the nodes whose characteristics significantly deviate from the majority of other nodes in terms of structure and attributes. More specifically, we aim to give a numeric score $\mathit{y_i} \in [0, 1]$ for each node $\mathit{v_i}$ such that a node with high $\mathit{y_i}$ is deemed anomalous. Note that as an unsupervised method we do not have access to nodes' anomaly labels, but we do have information on a portion of nodes' class labels, where $\mathit{C}$ denotes the number of classes.

\begin{figure}[t]
\includegraphics[width=0.85\linewidth]{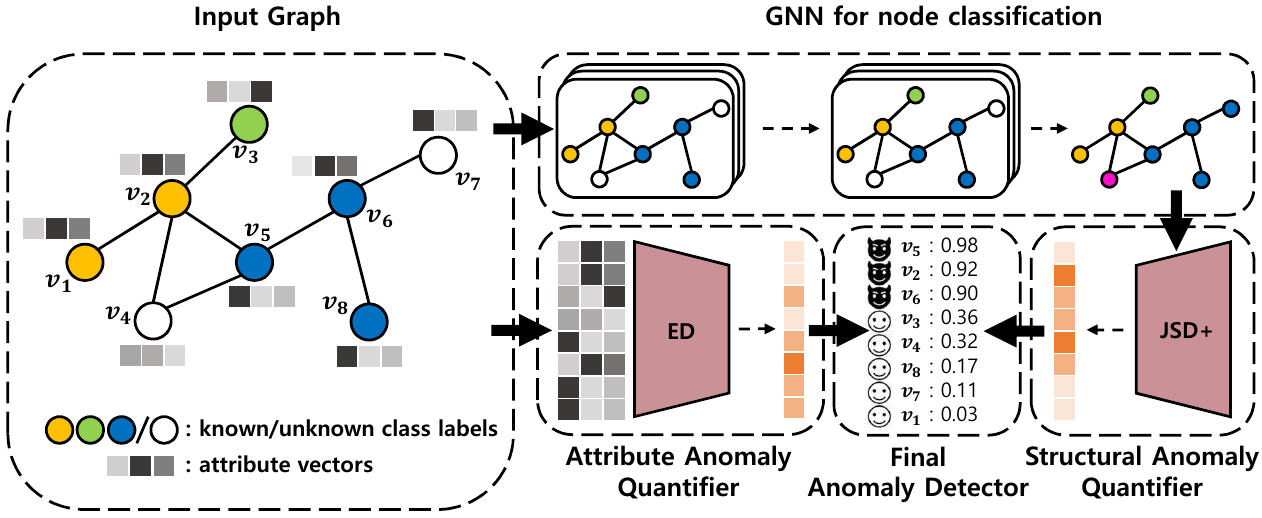}
\vspace{-3ex}
\caption{Overall framework of CLAD}
\vspace{-2ex}
\label{fig:architecture}
\end{figure}

\section{Method}
Fig.~\ref{fig:architecture} illustrates the overall architecture of CLAD. First, a {GNN for node classification} is trained in a semi-supervised manner using the known class labels. Here, ground-truth class labels are used when available; when not, pseudo-labels obtained from simple clustering algorithms on the node attributes can be used. We then devise \textit{two modules to measure the structural and attribute anomality} of a node. The final anomaly score for each node is calculated as a \textit{weighted sum of the scores} from both modules.

\subsection{Structural Anomaly Quantifier} \label{method}
\begin{figure}[t]
\centering
\vspace*{-6mm}
\subfloat[]{
\includegraphics[width=3cm]{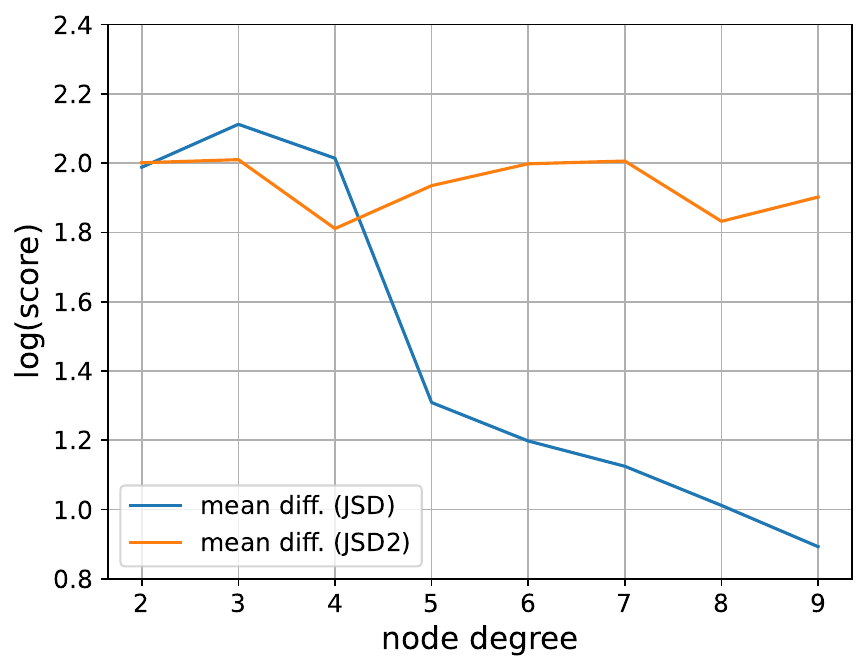}
\label{fig:method-a}
}
\subfloat[]{
\includegraphics[width=3cm]{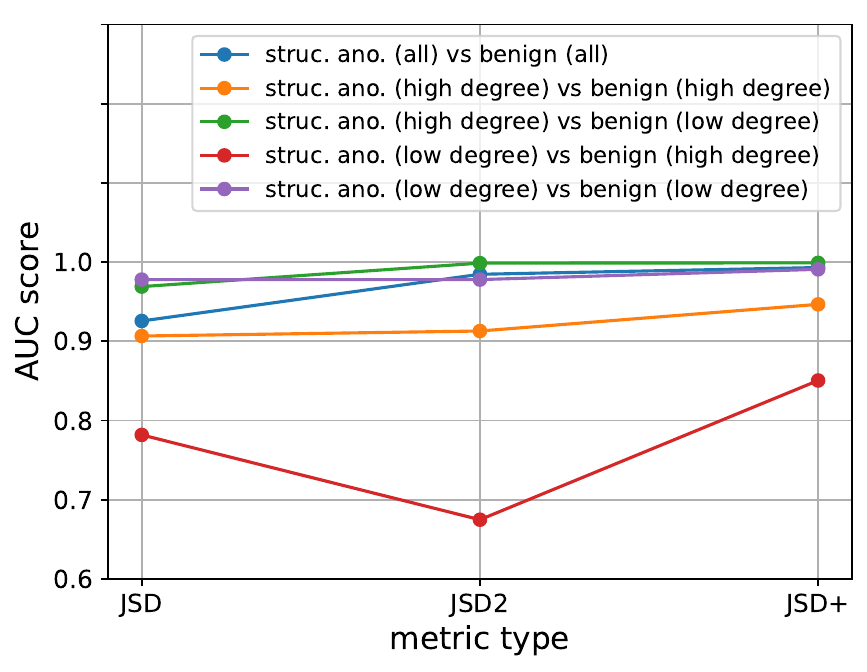}
\label{fig:method-b}
}
\\
\vspace*{-4mm}
\subfloat[]{
\includegraphics[height=1.5cm]{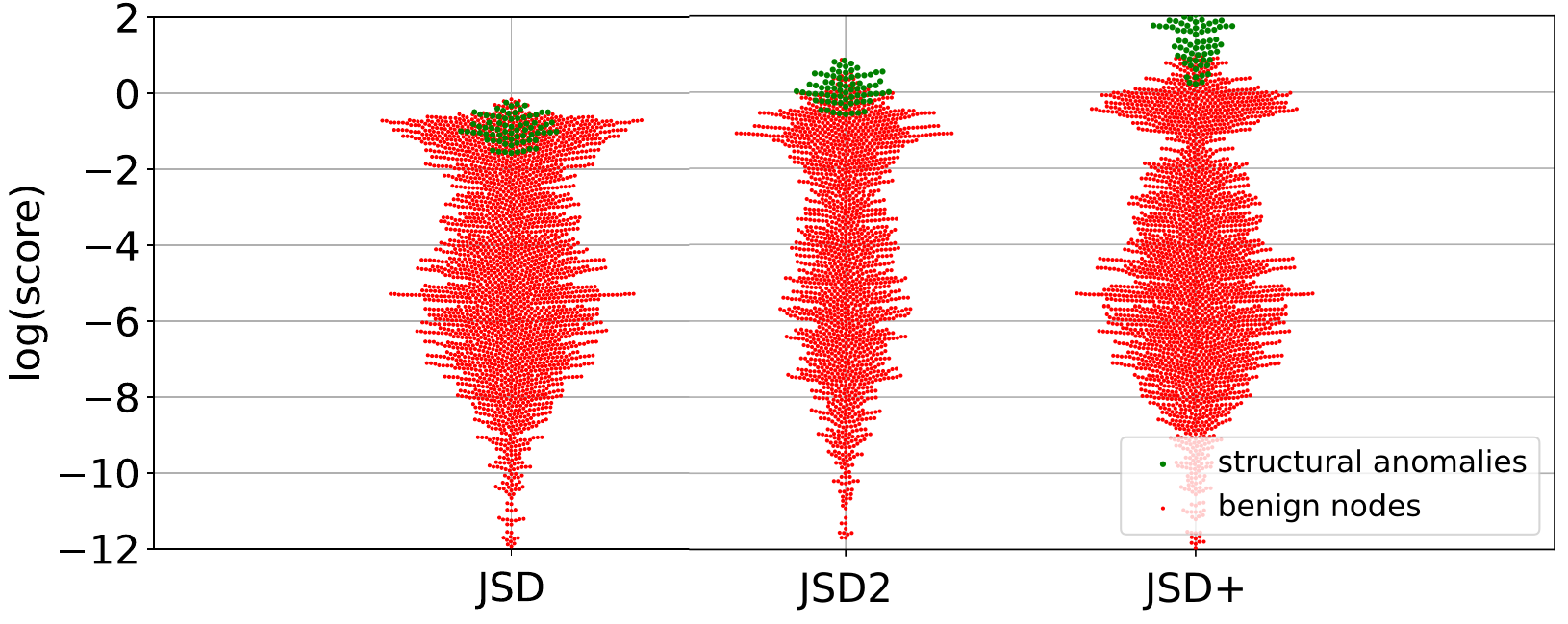}
\label{fig:method-c}
}
\subfloat[]{
\includegraphics[height=1.5cm]{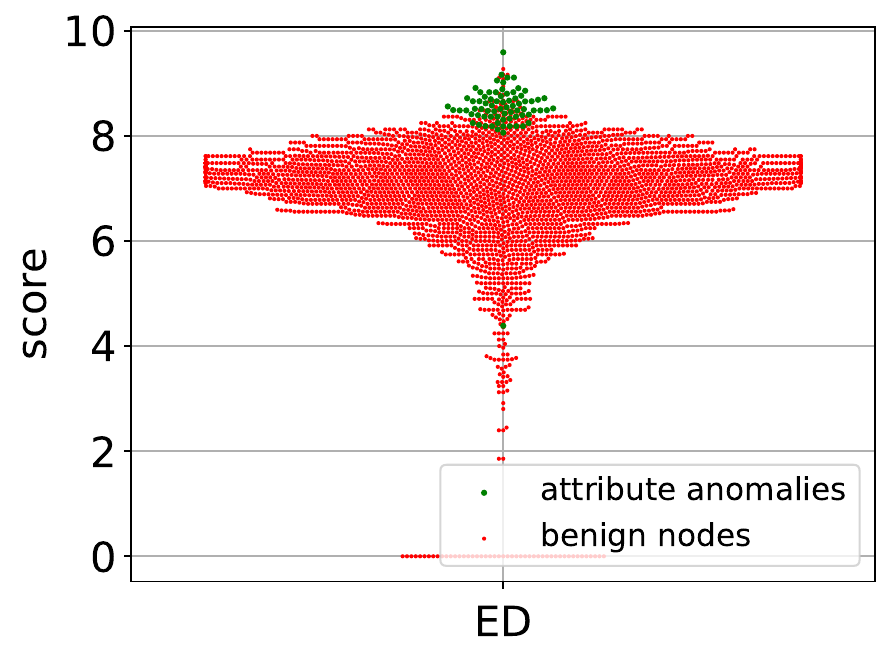}
\label{fig:method-d}
}
\vspace{-3ex}
\caption{(a) Mean difference of JSD and JSD2 scores between structural anomalies and benign nodes over node degrees. (b) ROC-AUC scores for JSD, JSD2 and JSD+ distinguishing nodes of different node degree groups. (c) Swarm plots of JSD, JSD2 and JSD+  scores for structural anomalies and benign nodes. (d) Swarm plot of ED scores for attribute anomalies and benign nodes. Citeseer dataset is used.}
\vspace{-5ex}
\label{fig:method}
\end{figure}

Jensen-Shannon Divergence (JSD)~\cite{jsd2} is a metric that measures the divergence of a set of probabilities, which is defined as:
\begin{equation}
\small
JSD(\mathit{i}) = \mathit{H}\left(\frac{1}{|\mathcal{N}(\mathit{i})|}  \sum_{\mathit{j} \in {\mathcal{N}(\mathit{i})}} \mathit{p_j}\right) - \frac{1}{|\mathcal{N}(\mathit{i})|}  \sum_{\mathit{j} \in {\mathcal{N}(\mathit{i})}} \mathit{H(p_j)}
\end{equation}
where $\mathit{p_j}\in\mathbb{R}^C$ is the output softmax probabilities of a GNN for node $\mathit{v_j}$. Regarding $\mathit{p_j}$ as a probability mass function over possible class labels,  $\mathit{H(p_j)} = - \sum_\mathit{c} \mathit{p_{jc}} \cdot log(\mathit{p_{jc}})$ is the Shannon entropy of $\mathit{p_j}$, where $\mathit{p_{jc}}$ is the probability of node $\mathit{v_j}$ belonging to class $\mathit{c} \in [1, 2, ..., C]$. $\mathcal{N}(\mathit{i})$ is the set of 1-hop neighboring nodes of a node $\mathit{v_i}$ including the node itself, i.e., $\mathit{v_i}$. 
In short, ${JSD(\mathit{i})}$ is increased when the discrepancy in the class distribution of node $\mathit{v_i}$ and its neighboring nodes $\mathcal{N}(i)$ becomes greater. From Fig.~\ref{fig:citeseer_entropy}, we can expect the JSD values of structural anomalies to be higher than those of benign nodes.
Moreover, the first plot of Fig.~\ref{fig:method-c} shows a fairly clear distinction in the $log(JSD)$ value between structural anomalies (in green) and benign nodes (in red), indicating that JSD can indeed be used as a metric for detecting structural anomalies.

\smallskip
\noindent{\textbf{Addressing node degree-bias issue in JSD. }}
However, naively utilizing JSD in its base form has one major drawback. That is,
\looseness=-1
when neighboring nodes are added/deleted, the JSD value of a relatively low-degree structural anomaly would be largely affected, while a high-degree structural anomaly would see a negligible change. This would make it difficult to differentiate high-degree structural anomalies and benign nodes, potentially hindering the effectiveness of JSD. To alleviate this node degree-bias issue, we consider the node degree in a way that maintains an appropriate difference in the values, which can be achieved by simply multiplying the node degree with the JSD value: $JSD2(\mathit{i}) = JSD(\mathit{i}) \cdot log(\mathit{degree_i})$. In Fig.~\ref{fig:method-a} we can see our intuition in action - for JSD the difference between structural anomalies and benign nodes sharply declines with an increase in the node degree, while for JSD2 anomalies maintain this difference. Moreover, Fig.~\ref{fig:method-c} shows that JSD2 better distinguishes structural anomalies from benign nodes.

Yet, while JSD2 alleviates the problem of \textit{difference between scores}, it runs into a new issue regarding the \textit{actual value of scores}. Specifically, by multiplying the node degree with the JSD value, we are at risk of obtaining an extremely large value for high-degree nodes regardless of their anomality, which is not desired. Consequently, JSD2 can potentially predict benign nodes to be anomalous if their node degree is high, and vice versa for low-degree structural anomalies. This is evident in Fig.~\ref{fig:method-b}, where we observe that JSD2 struggles more in distinguishing low-degree structural anomalies from high-degree benign nodes (as depicted by the red line).

\looseness=-1
This issue can be handled by using a value that doesn't become too large for high-degree benign nodes. Based on our observation in Fig.~\ref{fig:citeseer_entropy}, we can expect that a benign node's neighboring nodes are likely to share the same class label as the benign node itself, while the opposite holds for structural anomalies. Thus, defining the number of neighboring nodes whose predicted class is the same as node $v_i$ as $\gamma_{\mathit{i}}= \textstyle\sum_{\mathit{j} \in {\mathcal{N}(\mathit{i})}} (\arg\max_c \mathit{p_i} == \arg\max_c \mathit{p_j})$ (note that for obtaining $\gamma_{\mathit{i}}$, $v_i \notin \mathcal{N}(\mathit{i})$), the value $(\mathit{degree_i} - \gamma_{\mathit{i}})$ will tend to be small for benign nodes and large for structural anomalies.
Utilizing this value we devise a new metric: $JSD\texttt{+}(\mathit{i}) = JSD(\mathit{i}) \cdot log\left(\mathit{degree_i} - \gamma_{\mathit{i}} \right)$.
JSD+ addresses the aforementioned problem by giving less weight to the node degree of benign nodes. In Figs.~\ref{fig:method-b} and~\ref{fig:method-c}, we observe that JSD+ improves significantly over JSD and JSD2, especially in discriminating low-degree structural anomalies from high-degree benign nodes. Thus, our \textit{structural anomaly quantifier} uses JSD+ to represent the structural anomality of each node.

\subsection{Attribute Anomaly Quantifier}
To detect attribute anomalies, we compute the discrepancy of a node's attributes compared to its neighboring nodes' attributes. We first define the divergence between two nodes $\mathit{v_i}$ and $\mathit{v_j}$ as $dist(i, j) = \lVert \mathit{x_i} - \mathit{x_j}\rVert_2$ i.e., the Euclidean distance. Then, the discrepancy metric used for our \textit{attribute anomaly quantifier} is defined as: 
$ED(\mathit{i}) = \frac{1}{|\mathcal{N}(\mathit{i})|}  \sum_{\mathit{j} \in {\mathcal{N}(\mathit{i})}} dist(i, j)$,
where $\mathcal{N}(\mathit{i})$ is the set of 1-hop neighboring nodes of a node $\mathit{v_i}$, and we take the mean over the distances so that the number of neighbors does not affect the score. We expect the ED values to be high for attribute anomalies as their attributes would significantly differ from those of their neighbors. In Fig.~\ref{fig:method-d}, we observe that the ED values clearly distinguish attribute anomalies from benign nodes.

\smallskip
\noindent\textbf{Final Anomaly Detector. }
We compute the final anomaly score for node $\mathit{v_i}$ as a weighted sum of its structural and attribute anomaly scores~\cite{dominant, comga, anomalydae}: 
$\mathit{y_i} = \alpha \cdot struc_i + (1 - \alpha) \cdot attr_i$,
where $\mathit{struc_i}$ and $\mathit{attr_i}$ are the JSD+ and ED values obtained from the previous two modules scaled to be in range [0, 1], respectively, and $\alpha \in [0, 1]$ is a hyperparameter that balances the two scores. After ranking the final anomaly score for each node in descending order, we consider the nodes with high rank as anomalous.

\begin{table}[t]
\centering
\caption{Statistics of the datasets used for experiments}
\vspace{-3ex}
\resizebox{0.8\linewidth}{!}{
\begin{tabular}{|c|c c c c c|c|}
\hline
Dataset &\textit{N}&$\mathcal{|E|}$&\textit{F}&\# Anomalies&\# Labeled nodes&Class labels given?\\
\hline
Cora&2708&5803&1433&150&812 (30\%)&\multirow{5}{*}{$\bigcirc$}\\
Citeseer&3327&5077&3703&150&998 (30\%)&\\
Amazon Computers&13752&245861&767&750&4125 (30\%)&\\
Amazon Photo&7650&119081&745&450&2295 (30\%)&\\
ogbn-arxiv&169343&1249671&128&7500&50802 (30\%)&\\
\hline
Automotive&28922&106650&300&3173&\multirow{3}{*}{-}&\multirow{3}{*}{$\times$}\\
Patio, Lawn and Garden&30982&124740&300&3173&&\\
Office Products&47138&487848&300&3226&&\\
\hline
Yelp&45954&3846979&32&6677&\multirow{2}{*}{-}&\multirow{2}{*}{$\times$}\\
Elliptic&46564&73248&93&4545&&\\
\hline
\end{tabular}
}
\label{table:stats}
\vspace{-3ex}
\end{table}

\section{Experiments}

\begin{table*}[ht!]
\centering
\caption{AUC score (\%) of CLAD and baselines on all datasets. The computation time(s) is included in parantheses.}
\vspace{-2ex}
\resizebox{0.85\linewidth}{!}{
\begin{tabular}{|c|c c c c c|c c c|c c|}
\hline
&Cora&Citeseer&Amz. Com.&Amz. Pho.&ogbn-arxiv&Automotive&PL\&G&Office Products&Yelp&Elliptic\\
\hline
\hline
DOMINANT&89.6 (9.5)&87.4 (24.4)&63.4 (106.4)&64.3 (35.1)&77.9 (6651.7)&72.4 (350.9)&82.7 (407.2)&33.6 (904.5)&48.9 (740.4)&16.0 (790.2)\\
CoLA&89.9 (628.3)&89.7 (667.5)&68.5 (2513.3)&69.1 (1402.9)&80.3 (28704.6)&85.5 (4268.8)&86.0 (5131.4)&81.3 (7158.4)&43.6 (9314.6)&18.5 (6259.1)\\
ANEMONE&91.0 (256.4)&91.9 (542.2)&65.1 (2321.8)&65.9 (1184.7)&80.2 (24026.5)&60.9 4919.3)&66.9 (5094.6)&67.0 (7968.0)&41.8 (9503.1)&23.4 (6018.7)\\
AnomalyDAE&84.5 (10.1)&82.6 (27.2)&67.1 (116.8)&67.6 (33.2)&72.1 (7023.9)&77.5 (344.6)&74.1 (421.1)&76.4 (1046.5)&36.1 (767.0)&14.7 (819.6)\\
ComGA&92.2 (11.2)&90.9 (26.7)&69.7 (127.6)&70.1 (36.7)&81.0 (7203.5)&73.4 (367.0)&77.1 (448.6)&82.5 (943.8)&49.5 (786.3)&17.1 (847.9)\\
\hline
DOMINANT2&89.1 (17.4)&85.4 (17.6)&63.1 (32.7)&63.9 (20.9)&77.5 (1543.8)&71.1 (55.8)&78.6 (59.9)& 34.7 (146.7)&49.2 (190.3)&15.9 (175.8)\\
CoLA2&76.9 (467.2)&76.8 (933.7)&60.6 (3511.2)&59.8 (2238.0)&75.3 (40171.4)&75.3 (6515.2)&66.2 (6528.6)&66.4 (10432.3)&39.7 (12148.8)&18.7 (8041.0)\\
\hline
\textbf{CLAD (ours)}&\textbf{94.9} (3.5)&\textbf{97.3} (3.7)&\textbf{75.6} (28.1)&\textbf{79.0} (14.8)&\textbf{84.6} (1322.1)&\textbf{91.6} (47.2)&\textbf{89.8} (53.6)&\textbf{88.1} (139.5)&\textbf{56.3} (173.6)&\textbf{49.4} (50.2)\\
\hline
\end{tabular}
}
\label{table:aucresults}
\vspace{-2.3ex}
\end{table*}

\begin{table}[t]
\centering
\caption{AUC score (\%) of CLAD and baselines on three datasets. We report the results for detecting structural (s.) and attribute (a.) anomalies separately.}
\vspace{-2ex}
\resizebox{1\columnwidth}{!}{
\begin{tabular}{|c|c c|c c|c c|c c|c c|c c|c c|c c|}
\hline
&\multicolumn{2}{|c|}{DOMINANT}&\multicolumn{2}{|c|}{CoLA}&\multicolumn{2}{|c|}{ANEMONE}&\multicolumn{2}{|c|}{AnomalyDAE}&\multicolumn{2}{|c|}{ComGA}&\multicolumn{2}{|c|}{DOMINANT2}&\multicolumn{2}{|c|}{CoLA2}&\multicolumn{2}{|c|}{\textbf{CLAD (ours)}}\\
&s.&a.&s.&a.&s.&a.&s.&a.&s.&a.&s.&a.&s.&a.&s.&a.\\
\hline
Citeseer&77.0&95.8&93.2&80.9&96.4&87.5&88.2&81.5&94.3&82.6&73.9&95.2&87.2&65.2&\textbf{99.3}&\textbf{97.7}\\
Amz. Com.&69.8&55.9&79.2&54.6&73.0&50.2&78.8&52.0&80.3&51.5&69.9&55.6&68.1&50.2&\textbf{93.1}&\textbf{64.9}\\
Amz. Pho.&69.4&58.3&78.1&57.9&72.5&54.4&76.9&58.1&79.4&56.0&69.0&58.0&65.3&51.7&\textbf{96.5}&\textbf{68.7}\\
\hline
\end{tabular}
}
\label{table:bothresults}
\vspace{-2ex}
\end{table}

\begin{table}[t]
\centering
\small
\caption{AUC score (\%) of CLAD on three datasets while reducing the ratio/number of known class labels.}
\vspace{-2ex}
\resizebox{0.65\columnwidth}{!}{
\begin{tabular}{|c||c c c c|c c|}
\hline
&30\%&10\%&5\%&1\%&10&1\\
\hline
Citeseer&97.3&96.6&96.4&94.1&90.2&86.4\\
Amz. Com.&75.6&74.9&74.9&74.6&71.9&58.8\\
Amz. Pho.&79.0&78.9&78.8&78.8&70.7&57.3\\
\hline
\end{tabular}
}
\label{table:ratioresults}
\vspace{-3ex}
\end{table}

\noindent{\textbf{Datasets. }} We conduct extensive experiments on ten widely used datasets, which can be divided into three categories. (1) \textit{\textbf{Synthetic1}}: We use five benchmark datasets~\cite{citationdatasets, copurchasedatasets, ogbn} in which anomalies are added following a previous well-known anomaly injection scheme~\cite{dominant} (i.e., Cora, Citeseer, Amazon Computers, Amazon Photo, and ogbn-arxiv). 
(2) \textit{\textbf{Synthetic2}}: We create three datasets in which nodes are users in Amazon, edges denote the co-review relationship between the users, and node attributes are the pre-trained embeddings of the reviews written by each user. We assume that an anomalous user writes fake reviews to random items, and this leads to anomalous links in our generated datasets (i.e., Automotive, Patio,Lawn and Garden, and Office Products).
\textit{Synthetic2} datasets' anomalies follow a more realistic scenario than that of \textit{Synthetic1} datasets. 
(3) \textit{\textbf{Real-world}}: We use two datasets~\cite{yelp, elliptic} that contain real-world anomalies (i.e., Yelp and Elliptic).
Data statistics are summarized in Table~\ref{table:stats}. Note that ground-truth class labels are only present for \textit{Synthetic1} datasets.

\smallskip
\looseness=-1
\noindent{\textbf{Baselines. }} We compare CLAD with the following methods: DOMINANT~\cite{dominant}, CoLA~\cite{cola}, ANEMONE~\cite{anemone}, AnomalyDAE~\cite{anomalydae} and ComGA~\cite{comga}. We additionally include variants of DOMINANT and COLA, i.e., DOMINANT2 and CoLA2, by adding a cross-entropy loss for the node classification task to their respective objective functions. For all baselines, we use the official codes published by the authors, including the hyperparameter settings.

\smallskip
\noindent{\textbf{Implementation Details.}} We use a two-layer GCN~\cite{gcn} as the backbone GNN. For \textit{Synthetic1} datasets, we randomly sample 30\% of the ground-truth class labels. For \textit{Synthetic2} and \textit{Real-world} datasets that do not contain ground-truth class label information, we assign a pseudo-label to each node via K-means clustering with 5 clusters based on the node attributes. We then disregard all class label information except for the 50 nodes closest to each cluster centroid, i.e., we only have knowledge on 250 class labels, 50 per class. For all datasets the backbone GNN classifier is trained on 95\% of these \textit{known class labels} and validated on the remaining 5\%. We find the best performing $\alpha$ from $[0.0, 0.1, 0.2, ... 0.9, 1.0]$. The average value of five independent runs is reported for all results. We adopt the ROC-AUC metric to evaluate the performance of anomaly detection.

\subsection{Performance Analysis}
\noindent{\textbf{General Performance. }} Table~\ref{table:aucresults} shows that CLAD outperforms all baselines on \textit{Synthetic1} datasets, proving the effectiveness of our method in detecting anomalies present in graph data. This is further supported by Table~\ref{table:bothresults}, which reports the detection performance in terms of structural and attribute anomalies separately. 
These results demonstrate the effectiveness of our proposed metrics.
By comparing with DOMINANT2 and CoLA2, we can observe that naively using the class label information falls short of improving the detection performance, further proving the soundness of our proposed method. As a further appeal, we emphasize that CLAD achieves superior performance even while consuming much less computational time(GCN training and calculating scores) than baselines, which is especially beneficial for massive datasets like ogbn-arxiv. This shows the practicality of CLAD in the real-world.

\smallskip
\noindent{\textbf{Reducing the ratio/number of known class labels. }} We further report the performance of CLAD for cases in which only a few nodes are labeled with class information. Table~\ref{table:ratioresults} shows that reducing the ratio down to 1\% only incrementally hinders the performance. Even with access to only a handful (e.g., 10) of class labels, CLAD produces results comparable to that of baselines. 

\smallskip
\noindent{\textbf{Lack of class labels. }} As previously mentioned, experiments on \textit{Synthetic2} and \textit{Real-world} datasets are conducted under the assumption that \textit{class labels are not available}. In Table~\ref{table:aucresults}, we observe that CLAD still outperforms all baselines even with pseudo-labels obtained based on node attributes. Note that CLAD shows significant performance gain especially in \textit{Real-world} datasets. We argue that these results demonstrate the practicality of CLAD in real-world scenarios where the class label information may be missing.

\begin{figure}[t]
\vspace{-2ex}
\centering
\subfloat[]{
\includegraphics[width=3.1cm]{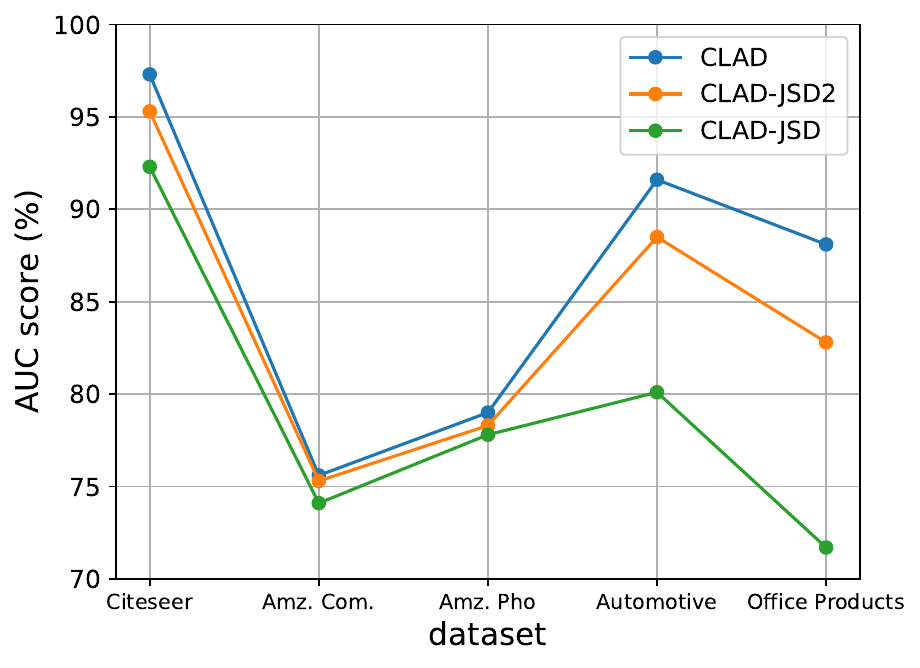}
\label{fig:experiment-a}
}
\subfloat[]{
\includegraphics[width=3.1cm]{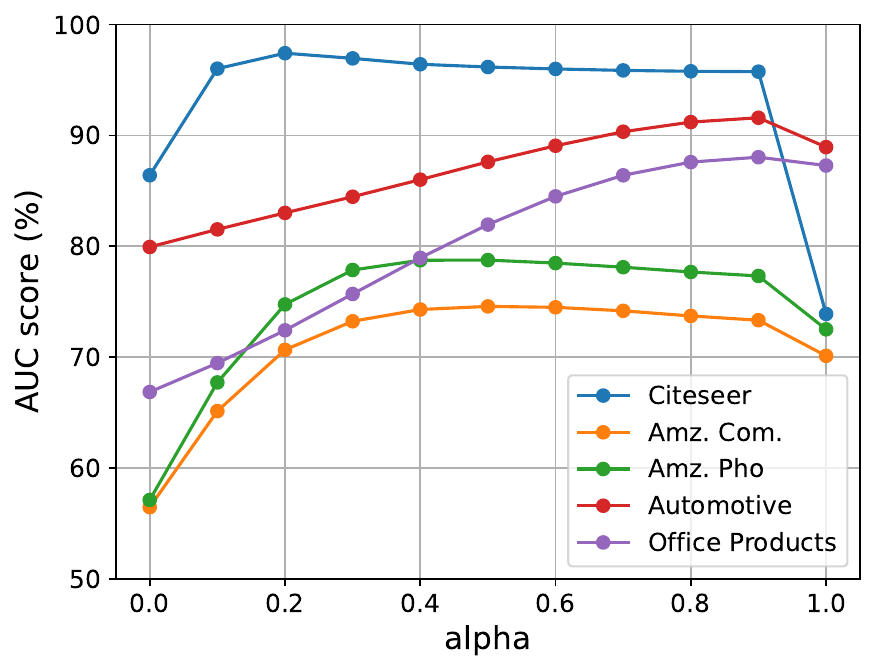}
\label{fig:experiment-b}
}
\vspace{-4ex}
\caption{(a) AUC scores (\%) for CLAD and its two variants. (b) Effect of parameter $\alpha$ on CLAD's detection performance.}
\label{fig:experiment}
\vspace{-4ex}
\end{figure}

\smallskip
\noindent{\textbf{Ablation Study. }} We compare the results of CLAD with two variants that replace $JSD+$ with $JSD$ and $JSD2$, respectively. Fig.~\ref{fig:experiment-a} shows that the original CLAD model outperforms both CLAD-JSD2 and CLAD-JSD. 
Since the output of the attribute anomaly quantifier does not change, the performance differences only stem from the choice of the metric in the \textit{structural anomaly quantifier}. This demonstrates the superiority of $JSD+$ over $JSD$ and $JSD2$.

\smallskip
\noindent{\textbf{Hyperparameter Sensitivity. }} We investigate the sensitivity of parameter $\alpha$. In Fig.~\ref{fig:experiment-b}, we observe that $\alpha$ should be set to a value between 0 and 1, indicating that jointly taking into account both perspectives of anomalies is important for achieving high detection performance. Moreover, an optimal $\alpha$ differs across datasets.

\section{Conclusion}
In this paper, we propose a Class Label-aware Graph Anomaly Detection framework (CLAD) that utilizes a limited amount of labeled nodes to enhance the performance of unsupervised GAD. Specifically, based on our empirical analysis that the vast majority of benign nodes have a much lower entropy of the neighborhood class distribution compared with that of structural anomalies, we devise a metric, called $JSD+$, that effectively and efficiently distinguishes structural anomalies from benign nodes.
Extensive experimental results demonstrate that CLAD outperforms existing unsupervised GAD methods, even in the absence of ground-truth class label information - indicating CLAD's versatility in real-world scenarios.

\smallskip
\noindent{\textbf{Acknowledgement. }} No.2022-0-00077 and No.2021R1C1C1009081.
\bibliographystyle{ACM-Reference-Format}
\bibliography{reference}

\end{document}